\documentclass[11pt]{article}
\usepackage{eacl2017}
\usepackage{times}
\usepackage{url}
\usepackage{latexsym}
\usepackage{multirow}
\usepackage{todonotes}

\eaclfinalcopy % Uncomment this line for the final submission
\def\eaclpaperid{6} %  Enter the acl Paper ID here

\title{Replication issues in syntax-based aspect extraction for opinion mining}

\author{Edison Marrese-Taylor, Yutaka Matsuo \\
  Department of Technology Management for Innovation\\
  The University of Tokyo\\
  7-3-1 Hongo, Bunkyo-ku, Tokyo 113-8656, Japan\\
  {\tt emarrese,matsuo@weblab.t.u-tokyo.ac.jp} \\}

\date{}

\begin{document}

\maketitle

\begin{abstract}
Reproducing experiments is an important instrument to validate previous work and build upon existing approaches. It has been tackled numerous times in different areas of science. In this paper, we introduce an empirical replicability study of three well-known algorithms for syntactic centric aspect-based opinion mining. We show that reproducing results continues to be a difficult endeavor, mainly due to the lack of details regarding preprocessing and parameter setting, as well as due to the absence of available implementations that clarify these details. We consider these are important threats to validity of the research on the field, specifically when compared to other problems in NLP where public datasets and code availability are critical validity components. We conclude by encouraging code-based research, which we think has a key role in helping researchers to understand the meaning of the state-of-the-art better and to generate continuous advances.
\end{abstract}

\section{Introduction}

Aspect-based opinion mining is one of the main frameworks for sentiment analysis. It aims to extract fine-grained opinion targets from opinion texts and its importance resides in the fact that without knowing the aspects, opinion analyses are of limited use \cite{liu_sentiment_2012}. The concept originated more than 10 years ago as a specific case of sentiment analysis and has gradually gained relevance as a concrete and complete problem in opinion mining. The key task in aspect-based sentiment analysis is to extract the aspects or targets that have been commented in opinion documents. Sentiment orientation can be obtained later based on the extracted terms using or adapting any of the generic approaches for sentiment classification. Therefore, an important amount of focus has been posed on the problem of aspect extraction. 

Researchers have proposed several methods for aspect extraction so far, and many authors consider that these approaches largely fall into three main categories. On the one hand, we find syntactical or linguistic methods, which are generally based on other basic NLP tasks, such as POS-tagging and parsing, plus some fixed rules or rankings mechanisms. On the other hand, we find purely statistical approaches, which are mainly based on topic modeling. Finally, we also find extensive work on supervised learning methods, in which case the problem is approached as sequence labeling. Experiments with both Neural Networks and Graphical Models have reported fairly good results so far.

A review of the literature in syntactical approaches showed us that most of the proposed ideas are inspired by or directly built on top of previous methods. Papers generally include detailed comparisons of the approaches, but we found very few publications accompanied by code releases that make it easier to effectively compare and contrast methods. We believe the lack of code availability is increasingly becoming a threat to validity in the field by adding layers of obscurity to new approaches, specially to those that are built on top of previous ideas. 

Given the current state in the field, in this paper we study replicability issues in aspect-based opinion mining. We focus on syntactic methods, which tend to show a lower degree of transparency due to the increasing level of model complexity and the lack of code availability. In that sense, in this work we want to encourage discussion on this topic by addressing some key questions.

\begin{enumerate}
\item Are the explanations given in the papers generally sufficient to replicate the proposed models?
\item Do differences in preprocessing have a big impact on performance? 
\item Do parameters need to be heavily tuned in order to achieve the reported performance?
\end{enumerate}

Our goal throughout this paper is to start exploring possible answers to these questions and provide an environment for further discussions and improvements. We will try to tackle the questions keeping in mind that reproducibility of an experimental result is a fundamental assumption in science. As we will see in the next section, the inability to replicate the experimental results published in a paper is an issue that has been considered in various other machine learning and computer science conferences. There have been several discussions arising from this issue and there seems to be a widespread view that we need to do something to address this problem. We would like to join this quest too. 

\section{Related Work}

Aspect-based opinion mining aims to identify the aspects of entities being reviewed in an text and to determine the sentiment reviewers express for each aspect. Aspects usually correspond to arbitrary topics considered important or representative of the text that is being analyzed. 

The aspect-based approach has become fairly popular. Since its conception, arguably after \newcite{hu_mining_2004}, many unsupervised approaches based on statistical and syntax analysis such as \newcite{qiu_opinion_2011} and \newcite{liu_opinion_2012} have been developed. While here we specifically tackle these kind of models, other popular unsupervised techniques such as \newcite{mukherjee_aspect_2012} are based on LDA.

On the other hand, existing supervised approaches in the field are mainly based on sequence labeling. Since 2014 the SemEval workshop included a shared task on the topic \cite{pontiki_semeval-2014_2014}, which has also encouraged the development of new supervised methods. We find approaches based on CRFs such as \newcite{mitchell_open_2013} and deep learning \cite{irsoy_opinion_2014} \cite{liu_fine-grained_2015}, \cite{zhang_neural_2015}.

The replicability issue has been tackled numerous times in different areas of science. For example, \newcite{casadevall_reproducible_2010} explore the importance and limits of reproducibility in scientific manuscripts in the field of Microbiology. In the field of Machine Learning, \newcite{drummond_replicability_2009} discusses issues arising from the inability to replicate the experimental results published in a paper. Also, \newcite{raeder_consequences_2010} show that when comparing the performance of different techniques some methodological choices can have a significant impact on the conclusions of any study. 

Furthermore, we also find studies in Software Engineering. For example, \newcite{monperrus_critical_2014} aimed to contribute to the field with a clarification of the essential ideas behind automatic software repair and included an in-depth critical analysis of \newcite{kim_automatic_2013}, an approach that had been published the year before in the same conference.

It is also possible to find work on replicability in Natural Language Processing. Conferences such as CICLing have undertaken maximum effort ---though so far rather fruitless--- in order to address the topic, giving a special prize every year to the best replicable paper\footnote{http://cicling.org/why\_verify.htm}. In addition, the proceedings of the ACL conference have included words on this topic on several occasions. For example, \newcite{kilgarriff_googleology_2007} introduces the issues of data cleaning and pre-processing, specially for those cases that involve crawling and/or downloading linguistic data. The paper claims that even though expertise and tools are available for most of these preprocessing steps, such as lemmatizers and POS-taggers for many languages, in the middle there is a logjam and questions always arise. The authors indicate that it seems to be undeniable that even though cleaning is a low-level, unglamorous task, it is yet crucial: the better it is done, the better the outcomes. All further layers of linguistic processing depend on the cleanliness of the data.

On the other hand, \newcite{pedersen_empiricism_2008} presents the sad tale of the Zigglebottom tagger, a fictional tagger with spectacular results. However, the code is not available and a new implementation does not yield the same results. In the end, the newly implemented Zigglebottom tagger is not used, because it does not lead to the promised results and all effort went to waste. \newcite{fokkens_offspring_2013} go further and actually experiment with what they think may be a real-world case of the Zigglebottom tagger, particularly, with the NER approach by \newcite{freire_2012}. The reimplementation process involved choices about seemingly small details such as rounding to how many decimals, how to tokenize or how much data cleanup to perform. They also tried different parameter combinations for feature generation, but the algorithm never yielded the exact same results. Particularly, their best run of their first reproduction attempt achieved nearly a 20\% drop in F-measure on average. Other authors such as \newcite{dashtipour_multilingual_2016} have worked on the same issue but for the task of sentiment classification, being unable to replicate the results of several papers. Our work is directly related to these since here we also attempt to re-implement other approaches.

\section{Empirical Replication Study}

As a first step, we first devoted ourselves to creating a friendly environment for experimentation. The goals of developing this framework were the following. (a) To provide a public Python implementation of notable algorithms for aspect extraction in aspect-based opinion mining that to date lack available implementations, (b) To provide an implementation that is easy to extend and thus to allow researchers to build novel approaches based on the routines we provide, and (c) To increase transparency in the field by providing full details about preprocessing steps, parameter setting and model evaluation. We are publicly releasing our code in GitHub\footnote{\url{github.com/epochx/opminreplicability}}, so it will welcome bug fixes, extensions and peer validation of its contents.

Our framework is an object-oriented package that is centered on the representation of a sentence as a property-rich object. Likewise, sentences are composed of tokens, which represent words and other punctuation marks with their respective properties such as stems, POS-tags, IOB-tags for chunks and dependency relation tags, among others. We have also developed wrappers for some popular packages for NLP, concretely the Stanford CoreNLP and Senna. This allows us to easily experiment with different tokenizers, stemmers, POS taggers, chunkers and parsers.

Our package also includes a module for corpora management, which provides easy access to the set of linguistic resources needed. We include parsers for word lists such as stopwords, opinion lexicons and also for more complex data structures regarding aspect-based opinion mining. In particular, we work with the well-known \textit{Customer Review Dataset} \cite{hu_mining_summarizing_2004,hu_mining_2004} which became the de facto benchmark for evaluation in syntax-based aspect-based opinion mining. This is also a very important part of our environment. 

We also include a simple module devoted to model evaluation, which makes the evaluation process transparent. We see aspect extraction as an information retrieval problem and thus the evaluation is based on precision, recall and F1-score, using exact matching to define a correctly extracted aspect.

On top of the framework we built our implementations of three different aspect extraction techniques, which we selected based on the approach they are based on, their novelty and their importance in the community. As we mentioned earlier, since we limit our study to syntactic approaches, here we explicitly omit algorithms that are intensively based on Web sources ---or other private sources or datasets--- and also approaches that use topic models or supervised learning models for sequence labeling. We selected three different papers, \newcite{hu_mining_2004}, \newcite{qiu_opinion_2011}, \newcite{liu_opinion_2012}. In the subsections below, we proceed to comment on the reasons for each choice and give details on our implementations.

\subsection{Frequency-Based Algorithm (FBA)}

We first consider the aspect extraction algorithm by \newcite{hu_mining_2004}, which pioneered on the problem of aspect-based opinion mining. This work is still being considered as a baseline for comparison and contrast with new approaches by most of the work on syntactic approaches in the literature. Despite this, there seems to be no available implementation of this technique to the best of our knowledge. These were our main motivations to work with this technique.

The aspect extraction procedure is based on frequent itemset mining, which given a database of transactions and a minimum support threshold $min_{sup}$ extracts the set of all the itemsets appearing in at least $min_{sup}$ transactions ---an itemset is just an unordered set of items in a transaction. In this case, each transaction is built using the nouns and words in the noun phrases of a sentence. Later, stopword removal, stemming and fuzzy matching are applied to the transactions in order to reduce the term dimensionality and to deal with word misspellings. Authors do not mention which stemming algorithm they use, so we resort both the well-known Porter stemmer and the Stanford lemmatizer, which can be regarded as the standard choices. 

Regarding fuzzy matching, the approach uses \newcite{jokinen_two_1991}, but authors simply state that [\textit{... all the occurrences of ``autofocus'' are replaced with ``auto-focus''}]. This description was insufficient to give us a full notion of how the process is carried out, specially since arbitrary word replacements can have an important impact when extracting aspects based on their frequency.

Similarly to \newcite{de_amorim_effective_2013}, who proposed a clustering method for spell checking, here we use clustering with the Levenshtein distance ratio as similarity metric to group terms. We tried with different strategies of hierarchical clustering and, based on our exploratory experiments, we decided to use \textit{complete linkage} to extract flat clusters so that the original observations in each flat cluster have a maximum cophenetic distance given by a parameter $min_{sim}$. Finally, we represent each stem as a fixed single stem in its cluster, keeping an index back from each of the original unstemmed words to its cluster, so we are later able to recover the terms as they appeared originally.

Authors later proceeded to mine \textit{frequent} occurring phrases by running the association rule miner CBA \cite{liu_integrating_1998}, which is based on the Apriori algorithm. The paper indicates that the Apriori algorithm works in two steps, first finding all frequent itemsets to later generate association rules from the discovered itemsets, so authors state they only needed the first step and use the CBA library for this part. This seems reasonable since it is a known fact that it is very efficient to use frequent itemsets to generate association rules \cite{agrawal_1993}. They limited itemsets to have a maximum of three words as they believed that a product feature contained no more than that number of terms. For minimum support, they defined an itemset as \textit{frequent} if it appeared in more than 1\% of the review sentences. In our case, since the CBA library was never released, we resort to an open-source implementation of the Apriori algorithm for frequent itemset mining \cite{borgelt_frequent_2012,pudi_efficiency_2002,pudi_armor_2003}.

After itemset mining, two pruning steps are applied in order to get rid of the incorrect, uninteresting and redundant features. We implemented these pruning techniques closely following the details given in the paper. 

Finally, extracted aspects are used to extract infrequent features that might also be important. In order to do so, they used terms in a lexicon as pivots to extract those nearby nouns that the terms modify. To generate the list of opinion words, they extracted the nearby adjective that modifies each feature on each of the sentences in which it appears, using stemming and fuzzy matching to take care of word variants and misspellings. The paper states that \textit{``a nearby adjective refers to the adjacent adjective that modifies the noun/noun phrase that is a frequent feature''}. However, it is not clear how they really find these adjectives. In our implementation, we defined a distance window from the aspect position and extract all adjectives that appear within this window. The size of this window became another parameter of the model. Finally, to extract infrequent features, authors checked those sentences that contain no frequent features but one or more opinion words and then extracted the nearest noun/noun phrase. 

We try to keep parameter setting as close as possible to the values reported by the original paper, but for POS-tagging we use CoreNLP or Senna instead of NLProcessor 2000. To obtain flat noun phrases, we use the Penn Treebank II output generated by the Stanford Constituency Parser and apply the same Perl script\footnote{\url{http://ilk.uvt.nl/team/sabine/homepage/software.html}} used to generate the data for the CoNLL-2000 Shared Task.

\subsection{Dependency-Based Algorithm (DBA)}

Our second implemented model is Double Propagation \cite{qiu_opinion_2011}, an approach that is fundamentally based on dependency relations between words for both aspect/target and opinion word extraction. This paper pioneered on the usage of dependency grammars to extract terms by iteratively using a set of eight rules based on dep-relations and POS-tags. Basically, the process starts with a set of \textit{seed} opinion words whose orientation is already known. In general, this is a reasonable assumption since several opinion lexicons already exist in the literature. The seeds are firstly used to extract aspects, which are defined as nouns that are modified by the seeds. Aspects are later used to extract more opinion words indicated by adjectives, other aspects and so on. This iterative process that propagates the knowledge with the help of the rules ends when no more opinion words or aspects are extracted. 

In the original paper, the set of dependency relationships given by the MINIPAR parser \cite{lin_minipar_2003} is used to develop the word extraction rules. We actually could not find the binaries on-line since the official website\footnote{\url{https://webdocs.cs.ualberta.ca/~lindek/minipar.htm}} is down; other binaries found on the Web were corrupted and unusable. This convinced us that MINIPAR can be regarded as a rather outdated model, so we decided to use the Stanford Parser \newcite{manning_stanford_2014} instead, which is among state-of-the-art models in the field. Our choice is supported by the results of \newcite{liu_automated_2015}, who successfully work with Double Propagation based on the Stanford dependency parser. Since the Stanford dep-tags differ from the tagset used by MINIPAR, we use the equivalences defined in the aforementioned paper.

After the extraction steps, the approach proceeds to apply a clause pruning phase. For each clause on each sentence, if it has more than one aspect and these are not connected by a conjunction, only the most frequent one is kept. In the paper, authors simply state that they [\textit{``identify the boundary of a clause using MINIPAR'']} and do not explain how to determine if the aspects are connected by the conjunction. We identify clauses by finding the set of non-overlapping parse sub-trees with label ``S''. To determine if the aspects are connected by any existing conjunction in a sentence, we simply check if the conjunction appears between the aspects in the same clause. 

The next step was to prune aspects that may be names of other products or names of product dealers, which may appear due to comparisons. In this case, the procedure is based on pre-defined patterns which are first matched in the text to later check if nearby nouns had previously been extracted as aspects. These are removed. The definition of \textit{nearby noun} is not given in the paper, so we add it as another parameter for the model, again using the notion of distance windows.

Finally, a rule is proposed to identify aspect phrases by combining each aspect with $Q$ consecutive nouns right before and after the aspect and $K$ adjectives before it. After obtaining the aspect phrases, another frequency-based pruning is conducted, removing aspects that appear only once in the dataset. Again, here we tried to set all the parameters as reported by the authors. Based on preliminary experiments, we decided to  also eliminate those terms that were extracted by leveraging on aspects that were later pruned, since they may introduce noise. 

\subsection{Translation-based Algorithm (TBA)}

The work of \newcite{liu_opinion_2012} is a novel application of classic statistical translation models and Graph Theory to extract opinion targets. Novelty and the good results obtained by the approach were our main motivations to work with this paper. 

For target extraction, the authors proposed a technique based on statistical word alignment. Specifically, they used a constrained version of the well-known IBM statistical alignment models \cite{brown_mathematics_1993}. The proposal is directly related to monolingual alignment, as proposed by \newcite{liu_collocation_2009}. For monolingual alignment, the parallel corpora fed to the model is simply two copies of the same corpus. At the same time, the condition that words cannot be aligned to themselves is added. \newcite{liu_opinion_2012} still use a monolingual parallel corpus but set the constraint that nouns and noun phrases can only be aligned to adjectives or vice-versa, meanwhile the rest of the words can be aligned freely. As a result, authors are able to capture noun/noun phrase-adjective relations that have longer spans that direct dependency relations in a sentence.

Since the IBM alignment models work at word granularity and then need to receive tokenized sentences as input, here we assume that authors first proceeded to group noun phrases in single tokens. According to the paper, they resorted to the \textit{C-value} technique \cite{frantzi_automatic_2000} for multi-word term extraction, which was originally developed to detect technical terminology in medical documents, but was also previously used in the domain of opinion mining by \newcite{zhu_multi_aspect_2009}. The method firstly generates a list of all possible multi-word terms and later ranks them using statistical features from the corpus. Even though in the original paper candidate multi-word terms are extracted using fixed patterns, authors decided to generate all candidates as simple n-grams (with $max_n=4$). We implemented and experimented with both fixed pattern and the n-gram versions for the \textit{C-value} technique. We also added a simple heuristic that works without ranking, grouping sets of nouns and other related figures that appear on the same parse NP sub-tree.

After the most likely constrained alignments are obtained for each sentence, authors estimated noun/noun phrase-adjective pair associations as the harmonic mean between the mutual translation probabilities. Finally, they built a bipartite graph with the words and estimate the confidence of each target candidate being a real target using the mined associations and applying a graph-based algorithm based on random walking. This is basically an iterative algorithm that exploits the mutual reinforcement between terms as given by the associations. Authors set the relevance of each target as the initial value of confidence, defining relevance as the normalized \textit{tf-idf} scores of the candidates, where \textit{tf} is the frequency on each term in the corpus and \textit{idf} is the inverse document frequency obtained using the Google N-gram Corpus.

In the paper, authors experimented with IBM1-3 models and showed that fertility parameters introduced by the third  model help to improve the performance by a small margin. The estimation of this model is rather complicated, in this case specially since it also includes a constrained version of the hill-climbing heuristic, so in our implementation we only include our versions of the IBM1-2 models. Regarding parameters, we set the proportion of candidate importance $\lambda=0.3$ and the maximum series power parameter $k=100$, as given by the original paper. To compute the initial relevance of each candidate, authors use the Google Ngram corpus to obtain the \textit{idf} of a term. Due to the lack of explanations on what they consider as a document, we resorted to the the English Wikipedia. To calculate the \textit{idf} score of a term, we count the number of articles that contain the queried term and compare it to the total number of articles. When we find no articles for a given term, we simple use a minimum article count of 1.

Finally, the authors stated that the targets with higher confidence scores than a certain threshold $t$ are extracted as the opinion targets, but they do not specify the value they use. We let our implementation output the unfiltered list of candidates and their confidences and find the best value of the threshold later.

\section{Preliminary Results}

As we have shown, the implementation process involved choices about several details that were not clear or not mentioned on the papers. In our experiments we have found that even when trying different parameter combinations we remain unable to yield the exact same results in the original papers. Below we summarize our best results and findings for each algorithm.

\begin{table}[h!]
\centering
\footnotesize
\begin{tabular}{|c|c|c|c|c|}
\hline
\multirow{2}{*}{\textbf{Corpus}} & \multicolumn{2}{c|}{\textbf{Original}} & \multicolumn{2}{c|}{\textbf{Ours}} \\ \cline{2-5} 
                                 & \textbf{P}         & \textbf{R}        & \textbf{P}       & \textbf{R}      \\ \hline
\textbf{Apex DVD Player}         & 0.797              & 0.743             & 0.389            & 0.355           \\ \hline
\textbf{Creative MP3 Player}     & 0.818              & 0.692             & 0.293            & 0.319           \\ \hline
\textbf{Nikon Camera}            & 0.792              & 0.710             & 0.265            & 0.457           \\ \hline
\textbf{Nokia Phone}             & 0.761              & 0.718             & 0.328            & 0.489           \\ \hline
\textbf{Canon Camera}            & 0.822              & 0.747             & 0.352            & 0.286           \\ \hline
\textbf{Average}                 & 0.8                & 0.72              & 0.325            & 0.381           \\ \hline
\end{tabular}
\caption{Performance comparison for FBA.}
\label{table:comparisonFB}
\end{table}

Table \ref{table:comparisonFB} compares our implementation's best results so far with the values reported by \newcite{hu_mining_2004}. We remain unable to replicate the performance reported by the authors and see a big drop for both precision and recall in all the datasets. In our experiments, we noted that the most sensitive parameter was $min_{sup}$ for itemset mining. We also experimented omitting the pruning steps and observed that precision and recall were not too different from the results we obtained with pruning.

We also observed that several parameters configurations conveyed the same final performance for each corpus. Among the 1470 per-corpus parameter configurations we tried, we found 18 optimal settings for both the Apex DVD Player and Canon Camera corpora, 16 for Nikon Camera, 6 for Creative MP3 Player and 3 for Nokia Phone. 

Differences in preprocessing did not offer consistent differences in performance. For the Apex DVD Player, Creative MP3 Player and Canon Camera corpora we found that processing with SennaConstParser + CoreNLPDepParser conveys the best results. For the Nikon Camera corpus, adding the PorterStemmer to the latter gave us the best performance. For the case of the  Nokia Phone corpus, the pipeline CoreNLPDepParser + CoNLL2000Chunker gave us the best results.

In the original paper, authors reported the performance of the model at different stages, showing that average values of precision an recall for the itemset mining stage are 0.68 and 0.56 respectively. We were surprised to find out that we could not even replicate these results, specially considering that only two parameters are at play at this level. As shown by the original paper, the final performance achieved is actually mainly due to the output of the itemset mining phase. We believe this is the reason why we observed some parameters have minimum impact on the performance. This means that no matter how good the pruning strategies are, results will not be as good as the original if we remain unable to replicate the output of the itemset mining phase.

\begin{table}[h!]
\centering
\footnotesize
\begin{tabular}{|c|c|c|c|c|}
\hline
\multirow{2}{*}{\textbf{Corpus}} & \multicolumn{2}{c|}{\textbf{Original}} & \multicolumn{2}{c|}{\textbf{Ours}} \\ \cline{2-5} 
                                 & \textbf{P}         & \textbf{R}        & \textbf{P}         & \textbf{R}         \\ \hline
\textbf{Apex DVD Player}         & 0.90               & 0.86              & 0.239              & 0.328              \\ \hline
\textbf{Creative MP3 Player}     & 0.81               & 0.84              & 0.180              & 0.319              \\ \hline
\textbf{Nikon Camera}            & 0.87               & 0.81              & 0.194              & 0.287              \\ \hline
\textbf{Nokia Phone}             & 0.92               & 0.86              & 0.287              & 0.359              \\ \hline
\textbf{Canon Camera}            & 0.90               & 0.81              & 0.201              & 0.356              \\ \hline
\textbf{Average}                 & 0.88               & 0.83              & 0.220              & 0.330              \\ \hline
\end{tabular}
\caption{Performance comparison for DBA.}
\label{table:comparisonDB}
\end{table}

Regarding DBA, Table \ref{table:comparisonDB} summarizes the results we obtained. Again, we see huge differences between our results and the ones reported by the original paper. Moreover, in this case we observe particularly low values for precision. A detailed review of the extracted aspects showed us that in fact many of the extracted terms do not correspond to aspects but rather to common nouns that are not related to the product. 

During experimentation, we also added support for different matching strategies ---for example, using word stems and including fuzzy matching as in FBA--- and although we observed improvements on the results, these were marginal. We used different opinion word seeds, firstly based only on the words ``good'' and ``bad'' and later using 9 same-size subsets of the opinion lexicon provided by Liu\footnote{http://www.cs.uic.edu/$\sim$liub/FBS/opinion-lexicon-English.rar}. In all cases, our best performing model uses one of these subsets.

As in the previous case, different parameter configurations led to the same performance for each corpus. In this case, among 240 parameter settings for each corpus, we found 12 optimal configurations for the Apex DVD Player corpus and 24 for each the other corpora. Regarding pre-processing, we could not use CoreNLP to transform the constituent trees given by Senna into dep-trees. Constituency trees seemed to be malformed and raised grammar parsing errors, therefore we only experimented using the CoreNLPDepParser + CoNLL2000Chunker pipeline.

\begin{table}[h!]
\centering
\footnotesize
\begin{tabular}{|c|c|}
\hline
\textbf{Corpus}				& $t^*$ \\ \hline
\textbf{Apex DVD Player}		& 160 	\\ \hline
\textbf{Creative MP3 Player}	& 200 	\\ \hline
\textbf{Nikon Camera}			& 100 	\\ \hline
\textbf{Nokia Phone}			& 90  	\\ \hline
\textbf{Canon Camera}			& 110 	\\ \hline
\end{tabular}
\caption{Optimal value of $t$ for each corpus.}
\label{table:best_t}
\end{table}

Table \ref{table:comparisonTB} shows a comparison of the results we have obtained so far using our implementation of TBA and the values provided by \newcite{liu_opinion_2012}. Once more, we remain unable to replicate the performance reported by the paper. 

On our experiments we tried with all three grouping strategies to generate multi-word terms; namely, our simple heuristic and \textit{C-value} using both n-grams and fixed patterns. We also tried adding a limit for the number of groups generated by the \textit{C-value} technique and used stemming to improve frequency counts. The ``ngram'' technique turned out to be the best performing on each corpus, although the limit parameter varies from case to case.

As we mentioned earlier, to evaluate the impact of the $t$ parameter whose value was not reported by \newcite{liu_opinion_2012}, we let the model return the unfiltered aspect candidates and evaluate the performance for $t \in [10, 20, ... ,t_{max}]$. Note that $t_{max}$ might be different each time. Because of this, the number of parameter configurations we tried for each corpora is slightly different. Instead of reporting each value, we rather report the average of per-corpus evaluated parameter settings, which was 1006. As Table \ref{table:best_t} shows, we found that rather than a single cross-corpus optimal value, this parameter needs to be tuned per-corpus. In this sense, when we experimented without setting a threshold we obtained a maximum recall of 0.697 ---for the Nokia Phone corpus--- but at the cost of precision 0.151. When we set $t=10$, we obtained a maximum precision of 0.9 but at the cost of recall being lower than five percent. These results mean the model does not seem to generalize well.

Since in our implementation we do not use the IBM3 model, we were aware we could see a difference in the performance. However, based on the results by the original paper, which showed that improvements of IBM3 over IBM2 are small ---about 5\%--- we think it is very unlikely this difference can explain the big drop in performance we have observed.

\begin{table}[h!]
\centering
\footnotesize
\begin{tabular}{|c|c|c|c|c|}
\hline
\multirow{2}{*}{\textbf{Corpus}} & \multicolumn{2}{c|}{\textbf{Original}} & \multicolumn{2}{c|}{\textbf{Ours}} \\ \cline{2-5} 
                                 & \textbf{P}         & \textbf{R}        & \textbf{P}       & \textbf{R}      \\ \hline
\textbf{Apex DVD Player}         & 0.89               & 0.87              & 0.362            & 0.389           \\ \hline
\textbf{Creative MP3 Player}     & 0.81               & 0.85              & 0.400            & 0.327           \\ \hline
\textbf{Nikon Camera}            & 0.84               & 0.85              & 0.380            & 0.404           \\ \hline
\textbf{Nokia Phone}             & 0.88               & 0.89              & 0.588            & 0.381           \\ \hline
\textbf{Canon Camera}            & 0.87               & 0.85              & 0.400            & 0.341           \\ \hline
\textbf{Average}                 & 0.86               & 0.86              & 0.426            & 0.368           \\ \hline
\end{tabular}
\caption{Performance comparison for TBA.}
\label{table:comparisonTB}
\end{table}

\section{Discussion and further directions}

The ongoing empirical study we introduce in this paper has provided concrete cases to help us answer the questions that motivate this paper. As seen, we have so far failed to reproduce the original results in the three studied cases. Even though several reasons may be the cause for this failure, we think further experimentation can allow us to determine the key elements that would explain the differences. In fact, our preliminary experiments have already helped us isolate specific parameters for each model that seem to more strongly improve the performance. Our results show that parameters that are closely related to the core of the extraction methods, such as $min_{sup}$ for FBA and the confidence threshold $t$ for TBA seem in fact to be playing these key roles. 

We are planning to run controlled experiments in order to isolate as much as possible the effect of each parameter or processing step and understand their interplay. This will enable us to tell where important implementation differences between our version and the original version may be. Given that we do not have access to the original codes, it is only by means of these inferred differences that we can gain real insights on where the keys for replicability lay.

We believe the results in this paper already prove that explanations given in the original papers were generally insufficient to correctly replicate the models. The lack of resources ---except for the evaluation datasets--- caused us to navigate in the dark as we could not reverse-engineer many intermediate steps. Certain details of preprocessing and parameter setting are only mentioned superficially or not at all in the original papers. However, many of these seemingly small details did make a big difference in our results. We understand there is often not enough space in the manuscripts to capture all details, specially since they are generally not the core of the research described. However, code releases play a critical role in uncovering these details and making research at least replicable. 

Regarding pre-processing, in our experiments so far with both Senna and CoreNLP we saw performance differences that are however not consistent, which seems to indicate that there is no optimal preprocessing pipeline for each algorithm. On the other hand, model parameters do not seem to be correlated with pre-processing choices, although we did find a single case in which a special pre-processing step lead to better results in a single corpus.

Though we could not replicate the results published in the original papers, we have shown that parameter values as reported by these papers do not necessarily yield the best results. Moreover, parameters that may seem unimportant turned out to cause important performance differences for us. Most parameters indeed had to be heavily tuned in order to achieve the best performance.

Finally, the poor results obtained by our implementations also leave us puzzled about how the evaluation is really performed on the original papers. Authors do not give much details on this topic. For example, \cite{hu_mining_2004} use stemming and simply eliminate some words from the text based on their fuzzy matching approach. This means their extracted terms are word stems only. However, we do not know if stemming is also applied to the gold standard to evaluate. We manually examined the \textit{Customer Review Dataset} and discovered that the manually extracted aspects do not seem to be stemmed. Moreover, we noted several inconsistencies in the annotation. This issue raises more questions for our research.

\section{Conclusions}

We have presented three replication cases in the domain of aspect-based opinion mining and shown that repeating experiments in the field is a complex issue. The experiments we designed and carried out have helped us answer our research questions, also raising some new ones. These answers seem to indicate that explanations on pre-processing, models specifications and parameter setting are generally insufficient to successfully replicate papers in the field. 

Our observations indicate that sharing data and software play key roles in allowing researchers to completely understand how methods work. Sharing is key to facilitating reuse, even if the code is imperfect and contains hacks and possibly bugs. Having access to such a set-up allows other researchers to validate research and to systematically test the approach in order to learn its limitations and strengths, ultimately allowing to improve on it.

\bibliography{eacl2017}

\begin{thebibliography}{}

\bibitem[\protect\citename{Agrawal \bgroup et al.\egroup }1993]{agrawal_1993}
Rakesh Agrawal, Tomasz Imieli\'{n}ski, and Arun Swami.
\newblock 1993.
\newblock Mining association rules between sets of items in large databases.
\newblock In {\em Proceedings of the 1993 ACM SIGMOD International Conference
  on Management of Data}, SIGMOD '93, pages 207--216, New York, NY, USA. ACM.

\bibitem[\protect\citename{Borgelt}2012]{borgelt_frequent_2012}
Christian Borgelt.
\newblock 2012.
\newblock Frequent item set mining.
\newblock {\em Wiley Interdisciplinary Reviews: Data Mining and Knowledge
  Discovery}, 2(6):437--456.

\bibitem[\protect\citename{Brown \bgroup et al.\egroup
  }1993]{brown_mathematics_1993}
Peter~F. Brown, Vincent J.~Della Pietra, Stephen A.~Della Pietra, and Robert~L.
  Mercer.
\newblock 1993.
\newblock The mathematics of statistical machine translation: {Parameter}
  estimation.
\newblock {\em Computational linguistics}, 19(2):263--311.

\bibitem[\protect\citename{Casadevall and
  Fang}2010]{casadevall_reproducible_2010}
A.~Casadevall and F.~C. Fang.
\newblock 2010.
\newblock Reproducible {Science}.
\newblock {\em Infection and Immunity}, 78(12):4972--4975, December.

\bibitem[\protect\citename{Dashtipour \bgroup et al.\egroup
  }2016]{dashtipour_multilingual_2016}
Kia Dashtipour, Soujanya Poria, Amir Hussain, Erik Cambria, Ahmad Y.~A.
  Hawalah, Alexander Gelbukh, and Qiang Zhou.
\newblock 2016.
\newblock Multilingual {Sentiment} {Analysis}: {State} of the {Art} and
  {Independent} {Comparison} of {Techniques}.
\newblock {\em Cognitive Computation}, 8(4):757--771.

\bibitem[\protect\citename{de Amorim and
  Zampieri}2013]{de_amorim_effective_2013}
Renato~Cordeiro de~Amorim and Marcos Zampieri.
\newblock 2013.
\newblock Effective {Spell} {Checking} {Methods} {Using} {Clustering}
  {Algorithms}.
\newblock In {\em Proceedings of the {International} {Conference} {Recent}
  {Advances} in {Natural} {Language} {Processing} {RANLP} 2013}, pages
  172--178, Hissar, Bulgaria, September. INCOMA Ltd. Shoumen, BULGARIA.

\bibitem[\protect\citename{Drummond}2009]{drummond_replicability_2009}
Chris Drummond.
\newblock 2009.
\newblock Replicability is not reproducibility: nor is it good science.
\newblock In {\em Proceedings of the {Evaluation} {Methods} for {Machine}
  {Learning} {Workshop}}, Montreal, Canada.

\bibitem[\protect\citename{Fokkens \bgroup et al.\egroup
  }2013]{fokkens_offspring_2013}
Antske Fokkens, Marieke van Erp, Marten Postma, Ted Pedersen, Piek Vossen, and
  Nuno Freire.
\newblock 2013.
\newblock Offspring from {Reproduction} {Problems}: {What} {Replication}
  {Failure} {Teaches} {Us}.
\newblock In {\em Proceedings of the 51st {Annual} {Meeting} of the
  {Association} for {Computational} {Linguistics} ({Volume} 1: {Long}
  {Papers})}, pages 1691--1701, Sofia, Bulgaria, August. Association for
  Computational Linguistics.

\bibitem[\protect\citename{Frantzi \bgroup et al.\egroup
  }2000]{frantzi_automatic_2000}
Katerina Frantzi, Sophia Ananiadou, and Hideki Mima.
\newblock 2000.
\newblock Automatic recognition of multi-word terms:. the {C}-value/{NC}-value
  method.
\newblock {\em International Journal on Digital Libraries}, 3(2):115--130.

\bibitem[\protect\citename{Freire \bgroup et al.\egroup }2012]{freire_2012}
Nuno Freire, Jos{\'e} Borbinha, and P{\'a}vel Calado, 2012.
\newblock {\em The Semantic Web: Research and Applications: 9th Extended
  Semantic Web Conference, ESWC 2012, Heraklion, Crete, Greece, May 27-31,
  2012. Proceedings}, chapter An Approach for Named Entity Recognition in
  Poorly Structured Data, pages 718--732.
\newblock Springer Berlin Heidelberg, Berlin, Heidelberg.

\bibitem[\protect\citename{Hu and Liu}2004a]{hu_mining_summarizing_2004}
Minqing Hu and Bing Liu.
\newblock 2004a.
\newblock Mining and {Summarizing} {Customer} {Reviews}.
\newblock In {\em Proceedings of the {Tenth} {ACM} {SIGKDD} {International}
  {Conference} on {Knowledge} {Discovery} and {Data} {Mining}}, {KDD} '04,
  pages 168--177, New York, NY, USA. ACM.

\bibitem[\protect\citename{Hu and Liu}2004b]{hu_mining_2004}
Minqing Hu and Bing Liu.
\newblock 2004b.
\newblock Mining {Opinion} {Features} in {Customer} {Reviews}.
\newblock In {\em Proceedings of the 19th {National} {Conference} on
  {Artifical} {Intelligence}}, {AAAI}'04, pages 755--760, San Jose, California.
  AAAI Press.

\bibitem[\protect\citename{Irsoy and Cardie}2014]{irsoy_opinion_2014}
Ozan Irsoy and Claire Cardie.
\newblock 2014.
\newblock Opinion {Mining} with {Deep} {Recurrent} {Neural} {Networks}.
\newblock In {\em Proceedings of the 2014 {Conference} on {Empirical} {Methods}
  in {Natural} {Language} {Processing} ({EMNLP})}, pages 720--728, Doha, Qatar,
  October. Association for Computational Linguistics.

\bibitem[\protect\citename{Jokinen and Ukkonen}1991]{jokinen_two_1991}
Petteri Jokinen and Esko Ukkonen.
\newblock 1991.
\newblock Two algorithms for approxmate string matching in static texts.
\newblock In {\em International {Symposium} on {Mathematical} {Foundations} of
  {Computer} {Science}}, pages 240--248. Springer.

\bibitem[\protect\citename{Kilgarriff}2007]{kilgarriff_googleology_2007}
Adam Kilgarriff.
\newblock 2007.
\newblock Googleology is bad science.
\newblock {\em Computational linguistics}, 33(1):147--151.

\bibitem[\protect\citename{Kim \bgroup et al.\egroup }2013]{kim_automatic_2013}
Dongsun Kim, Jaechang Nam, Jaewoo Song, and Sunghun Kim.
\newblock 2013.
\newblock Automatic patch generation learned from human-written patches.
\newblock In {\em Proceedings of the 2013 {International} {Conference} on
  {Software} {Engineering}}, pages 802--811. IEEE Press.

\bibitem[\protect\citename{Lin}2003]{lin_minipar_2003}
Dekang Lin.
\newblock 2003.
\newblock Dependency-{Based} {Evaluation} of {Minipar}.
\newblock In Anne Abeillé, editor, {\em Treebanks}, volume~20 of {\em Text,
  {Speech} and {Language} {Technology}}, pages 317--329. Springer Netherlands.

\bibitem[\protect\citename{Liu \bgroup et al.\egroup
  }1998]{liu_integrating_1998}
Bing Liu, Wynne Hsu, and Yiming Ma.
\newblock 1998.
\newblock Integrating classification and association rule mining.
\newblock In {\em Proceedings of the {Fourth} {International} {Conference} on
  {Knowledge} {Discovery} and {Data} {Mining}}.

\bibitem[\protect\citename{Liu \bgroup et al.\egroup
  }2009]{liu_collocation_2009}
Zhanyi Liu, Haifeng Wang, Hua Wu, and Sheng Li.
\newblock 2009.
\newblock Collocation extraction using monolingual word alignment method.
\newblock In {\em Proceedings of the 2009 {Conference} on {Empirical} {Methods}
  in {Natural} {Language} {Processing}: {Volume} 2-{Volume} 2}, pages 487--495.
  Association for Computational Linguistics.

\bibitem[\protect\citename{Liu \bgroup et al.\egroup }2012]{liu_opinion_2012}
Kang Liu, Liheng Xu, and Jun Zhao.
\newblock 2012.
\newblock Opinion {Target} {Extraction} {Using} {Word}-{Based} {Translation}
  {Model}.
\newblock In {\em Proceedings of the 2012 {Joint} {Conference} on {Empirical}
  {Methods} in {Natural} {Language} {Processing} and {Computational} {Natural}
  {Language} {Learning}}, pages 1346--1356, Jeju Island, Korea, July.
  Association for Computational Linguistics.

\bibitem[\protect\citename{Liu \bgroup et al.\egroup
  }2015a]{liu_fine-grained_2015}
Pengfei Liu, Shafiq Joty, and Helen Meng.
\newblock 2015a.
\newblock Fine-grained {Opinion} {Mining} with {Recurrent} {Neural} {Networks}
  and {Word} {Embeddings}.
\newblock In {\em Proceedings of the 2015 {Conference} on {Empirical} {Methods}
  in {Natural} {Language} {Processing}}, pages 1433--1443, Lisbon, Portugal,
  September. Association for Computational Linguistics.

\bibitem[\protect\citename{Liu \bgroup et al.\egroup
  }2015b]{liu_automated_2015}
Qian Liu, Zhiqiang Gao, Bing Liu, and Yuanlin Zhang.
\newblock 2015b.
\newblock Automated {Rule} {Selection} for {Aspect} {Extraction} in {Opinion}
  {Mining}.
\newblock In {\em Proceedings of the 24th {International} {Conference} on
  {Artificial} {Intelligence}}, {IJCAI}'15, pages 1291--1297, Buenos Aires,
  Argentina. AAAI Press.

\bibitem[\protect\citename{Liu}2012]{liu_sentiment_2012}
Bing Liu.
\newblock 2012.
\newblock Sentiment {Analysis} and {Opinion} {Mining}.
\newblock {\em Synthesis Lectures on Human Language Technologies}, 5(1):1--167.

\bibitem[\protect\citename{Manning \bgroup et al.\egroup
  }2014]{manning_stanford_2014}
Christopher~D. Manning, Mihai Surdeanu, John Bauer, Jenny Finkel, Steven~J.
  Bethard, and David McClosky.
\newblock 2014.
\newblock The {Stanford} {CoreNLP} {Natural} {Language} {Processing} {Toolkit}.
\newblock In {\em Association for {Computational} {Linguistics} ({ACL})
  {System} {Demonstrations}}, pages 55--60.

\bibitem[\protect\citename{Mitchell \bgroup et al.\egroup
  }2013]{mitchell_open_2013}
Margaret Mitchell, Jacqui Aguilar, Theresa Wilson, and Benjamin Van~Durme.
\newblock 2013.
\newblock Open {Domain} {Targeted} {Sentiment}.
\newblock In {\em Proceedings of the 2013 {Conference} on {Empirical} {Methods}
  in {Natural} {Language} {Processing}}, pages 1643--1654, Seattle, Washington,
  USA, October. Association for Computational Linguistics.

\bibitem[\protect\citename{Monperrus}2014]{monperrus_critical_2014}
Martin Monperrus.
\newblock 2014.
\newblock A critical review of automatic patch generation learned from
  human-written patches: essay on the problem statement and the evaluation of
  automatic software repair.
\newblock In {\em Proceedings of the 36th {International} {Conference} on
  {Software} {Engineering}}, pages 234--242. ACM.

\bibitem[\protect\citename{Mukherjee and Liu}2012]{mukherjee_aspect_2012}
Arjun Mukherjee and Bing Liu.
\newblock 2012.
\newblock Aspect {Extraction} {Through} {Semi}-supervised {Modeling}.
\newblock In {\em Proceedings of the 50th {Annual} {Meeting} of the
  {Association} for {Computational} {Linguistics}: {Long} {Papers} - {Volume}
  1}, {ACL} '12, pages 339--348, Stroudsburg, PA, USA. Association for
  Computational Linguistics.

\bibitem[\protect\citename{Pedersen}2008]{pedersen_empiricism_2008}
Ted Pedersen.
\newblock 2008.
\newblock Empiricism is not a matter of faith.
\newblock {\em Computational Linguistics}, 34(3):465--470.

\bibitem[\protect\citename{Pontiki \bgroup et al.\egroup
  }2014]{pontiki_semeval-2014_2014}
Maria Pontiki, Dimitris Galanis, John Pavlopoulos, Harris Papageorgiou, Ion
  Androutsopoulos, and Suresh Manandhar.
\newblock 2014.
\newblock {SemEval}-2014 {Task} 4: {Aspect} {Based} {Sentiment} {Analysis}.
\newblock In {\em Proceedings of the 8th {International} {Workshop} on
  {Semantic} {Evaluation} ({SemEval} 2014)}, pages 27--35, Dublin, Ireland,
  August. Association for Computational Linguistics and Dublin City University.

\bibitem[\protect\citename{Pudi and Haritsa}2002]{pudi_efficiency_2002}
Vikram Pudi and Jayant~R. Haritsa.
\newblock 2002.
\newblock On the {Efficiency} of {Association}-{Rule} {Mining} {Algorithms}.
\newblock In {\em Proceedings of the 6th {Pacific}-{Asia} {Conference} on
  {Advances} in {Knowledge} {Discovery} and {Data} {Mining}}, {PAKDD} '02,
  pages 80--91, London, UK, UK. Springer-Verlag.

\bibitem[\protect\citename{Pudi and Haritsa}2003]{pudi_armor_2003}
Vikram Pudi and Jayant~R. Haritsa.
\newblock 2003.
\newblock {ARMOR}: {Association} {Rule} {Mining} based on {ORacle}.
\newblock In {\em Proceedings of the {IEEE} {ICDM} {Workshop} on {Frequent}
  {Itemset} {Mining} {Implementations} ({FIMI}-03)}, volume~90, Melbourne,
  Florida, USA, November. CEUR Workshop Proceedings.

\bibitem[\protect\citename{Qiu \bgroup et al.\egroup }2011]{qiu_opinion_2011}
Guang Qiu, Bing Liu, Jiajun Bu, and Chun Chen.
\newblock 2011.
\newblock Opinion {Word} {Expansion} and {Target} {Extraction} {Through}
  {Double} {Propagation}.
\newblock {\em Computational Linguistics}, 37(1):9--27, March.

\bibitem[\protect\citename{Raeder \bgroup et al.\egroup
  }2010]{raeder_consequences_2010}
T.~Raeder, T.~R. Hoens, and N.~V. Chawla.
\newblock 2010.
\newblock Consequences of {Variability} in {Classifier} {Performance}
  {Estimates}.
\newblock In {\em 2010 {IEEE} {International} {Conference} on {Data} {Mining}},
  pages 421--430.

\bibitem[\protect\citename{Zhang \bgroup et al.\egroup
  }2015]{zhang_neural_2015}
Meishan Zhang, Yue Zhang, and Duy~Tin Vo.
\newblock 2015.
\newblock Neural {Networks} for {Open} {Domain} {Targeted} {Sentiment}.
\newblock In {\em Proceedings of the 2015 {Conference} on {Empirical} {Methods}
  in {Natural} {Language} {Processing}}, pages 612--621, Lisbon, Portugal,
  September. Association for Computational Linguistics.

\bibitem[\protect\citename{Zhu \bgroup et al.\egroup
  }2009]{zhu_multi_aspect_2009}
Jingbo Zhu, Huizhen Wang, Benjamin~K. Tsou, and Muhua Zhu.
\newblock 2009.
\newblock Multi-aspect opinion polling from textual reviews.
\newblock In {\em Proceedings of the 18th {ACM} conference on {Information} and
  knowledge management}, pages 1799--1802. ACM.

\end{thebibliography}
\bibliographystyle{eacl2017}

\appendix

\end{document}